\documentclass[runningheads]{llncs}
\usepackage[T1]{fontenc}
\usepackage{amsmath,booktabs,amssymb}
\usepackage{changepage}
\usepackage{multirow}
\usepackage[bb=boondox]{mathalfa}
\usepackage{subfig}
\usepackage{algcompatible}
\usepackage{amsmath}
\usepackage{adjustbox}
\usepackage[table,xcdraw]{xcolor}
\usepackage{dsfont}
\usepackage{booktabs}
\usepackage{multirow}
\usepackage{graphicx}
\usepackage[%
    colorlinks=true,
    pdfborder={0 0 0},
    linkcolor=blue
]{hyperref}
\hypersetup{
    colorlinks=true,
    linkcolor=blue,
    filecolor=magenta,      
    urlcolor=blue,
    citecolor=red,
}
\usepackage{paralist}
\usepackage{soul}
\usepackage{scalerel}
\usepackage{tikz}
\usetikzlibrary{svg.path}
\newcommand{\etal}{\textit{et al}.}
\begin{document}
\title{Semi-supervised Domain Adaptive Medical Image Segmentation through Consistency Regularized Disentangled Contrastive Learning}
\titlerunning{Disentangled Consistency Contrast}
%
\author{Hritam Basak, Zhaozheng Yin} 
%
\authorrunning{H Basak et al.}
%
\institute{Dept. of Computer Science, Stony Brook University, NY, USA}
%
\maketitle              
\begin{abstract}
Although unsupervised domain adaptation (UDA) is a promising direction to alleviate domain shift, they fall short of their supervised counterparts. In this work, we investigate relatively less explored semi-supervised domain adaptation (SSDA) for medical image segmentation, where access to a few labeled target samples can improve the adaptation performance substantially. Specifically, we propose a two-stage training process. First, an encoder is pre-trained in a self-learning paradigm using a novel domain-content disentangled contrastive learning (CL) along with a pixel-level feature consistency constraint. The proposed CL enforces the encoder to learn discriminative content-specific but domain-invariant semantics on a global scale from the source and target images, whereas consistency regularization enforces the mining of local pixel-level information by maintaining spatial sensitivity. This pre-trained encoder, along with a decoder, is further fine-tuned for the downstream task, (i.e. pixel-level segmentation) using a semi-supervised setting. Furthermore, we experimentally validate that our proposed method can easily be extended for UDA settings, adding to the superiority of the proposed strategy. Upon evaluation on two domain adaptive image segmentation tasks, our proposed method outperforms the SoTA methods, both in SSDA and UDA settings. Code is available at \href{https://github.com/hritam-98/GFDA-disentangled}{GitHub}.           

\keywords{Contrastive Learning  \and Style-content disentanglement \and Consistency Regularization \and Domain Adaptation \and Segmentation.}
\end{abstract}
\section{Introduction}\label{intro}
Despite their remarkable success in numerous tasks, deep learning models trained on a source domain face the challenges to generalize to a new target domain, especially for segmentation which requires dense pixel-level prediction. This is attributed to a {large semantic gap} between these two domains. Unsupervised Domain Adaptation (UDA) has lately been investigated to bridge this semantic gap between labeled source domain, and unlabeled target domain \cite{yao2022novel}, including adversarial learning for aligning latent representations \cite{xing2022low}, image translation networks \cite{yang2022source}, etc. However, these methods produce subpar performance because of the lack of supervision from the target domain and a large semantic gap in style and content information between the source and target domains. Moreover, when an image's \ul{content-specific information is entangled with its domain-specific style information, traditional UDA approaches fail to learn the correct representation of the domain-agnostic content while being distracted by the domain-specific styles.}
So, they cannot be generalized for multi-domain segmentation tasks \cite{chen2021semi}. 

Compared to UDA, obtaining annotation for a few target samples is worthwhile if it can substantially improve the performance by providing crucial target domain knowledge. Driven by this speculation, and the recent success of semi-supervised learning (SemiSL), we investigate {semi-supervised domain adaptation (SSDA)} as a potential solution.   
%
Recently, Liu \etal \cite{liu2022act} proposed an asymmetric co-training strategy between a SemiSL and UDA task, that complements each other for cross-domain knowledge distillation.  Xia \etal \cite{xia2020uncertainty} proposed a co-training strategy through pseudo-label refinement. Gu \etal \cite{gu2022contrastive} proposed a new SSDA paradigm using cross-domain contrastive learning (CL) and self-ensembling mean-teacher. However, these methods force the model to learn the low-level nuisance variability, which we know is insignificant to the task at hand. Hence, these methods fail to generalize if similar variational semantics are absent in the training set. Fourier Domain Adaptation (FDA) \cite{yang2020fda} was proposed to address these challenges by a simple yet effective spectral transfer method. Following \cite{yang2020fda}, we design a new Gaussian FDA to handle this cross-domain nuisance variability, without explicit feature alignment. 

Contrastive learning (CL) is another prospective direction where we enforce models to learn discriminative information from (dis)similarity learning in a latent subspace \cite{chaitanya2020contrastive,huy2022adversarial}. Liu \etal \cite{liu2022margin} proposed a margin-preserving constraint along with a self-paced CL framework, gradually increasing the training data difficulty. Gomariz \etal \cite{gomariz2022unsupervised} proposed a CL framework with an unconventional channel-wise aggregated projection head for inter-slice representation learning. However, traditional CL utilized for DA on images with entangled style and content leads to \ul{mixed representation learning, whereas ideally, it should learn discriminative content features invariant to style representation.}
Besides, the \ul{instance-level feature alignment of CL is subpar for segmentation, where dense pixel-wise predictions are indispensable} \cite{basak2023ideal}. 

To alleviate these three underlined shortcomings, we propose a novel contrastive learning with pixel-level consistency constraint via disentangling the style and content information from the joint distribution of source and target domain. Precisely, our contributions are as follows: \textbf{(1)} We propose to disentangle the style and content information in their compact embedding space using a joint-learning framework; \textbf{(2)} We propose encoder pre-training with two CL strategies: \textit{Style CL} and \textit{Content CL} that learns the style and content information respectively from the embedding space; \textbf{(3)} The proposed CL is complemented with a pixel-level consistency constraint with dense feature propagation module, where the former provides better categorization competence whereas the later enforces effective spatial sensitivity; 
\textbf{(4)} We experimentally validate that our SSDA method can be extended in the UDA setting easily, achieving superior performance as compared to the SoTA methods on two widely-used domain adaptive segmentation tasks, both in SSDA and UDA settings.  

\section{Proposed Method}\label{method}
\begin{figure}[t]
    \centering
    \includegraphics[width=0.95\columnwidth, height=70mm]{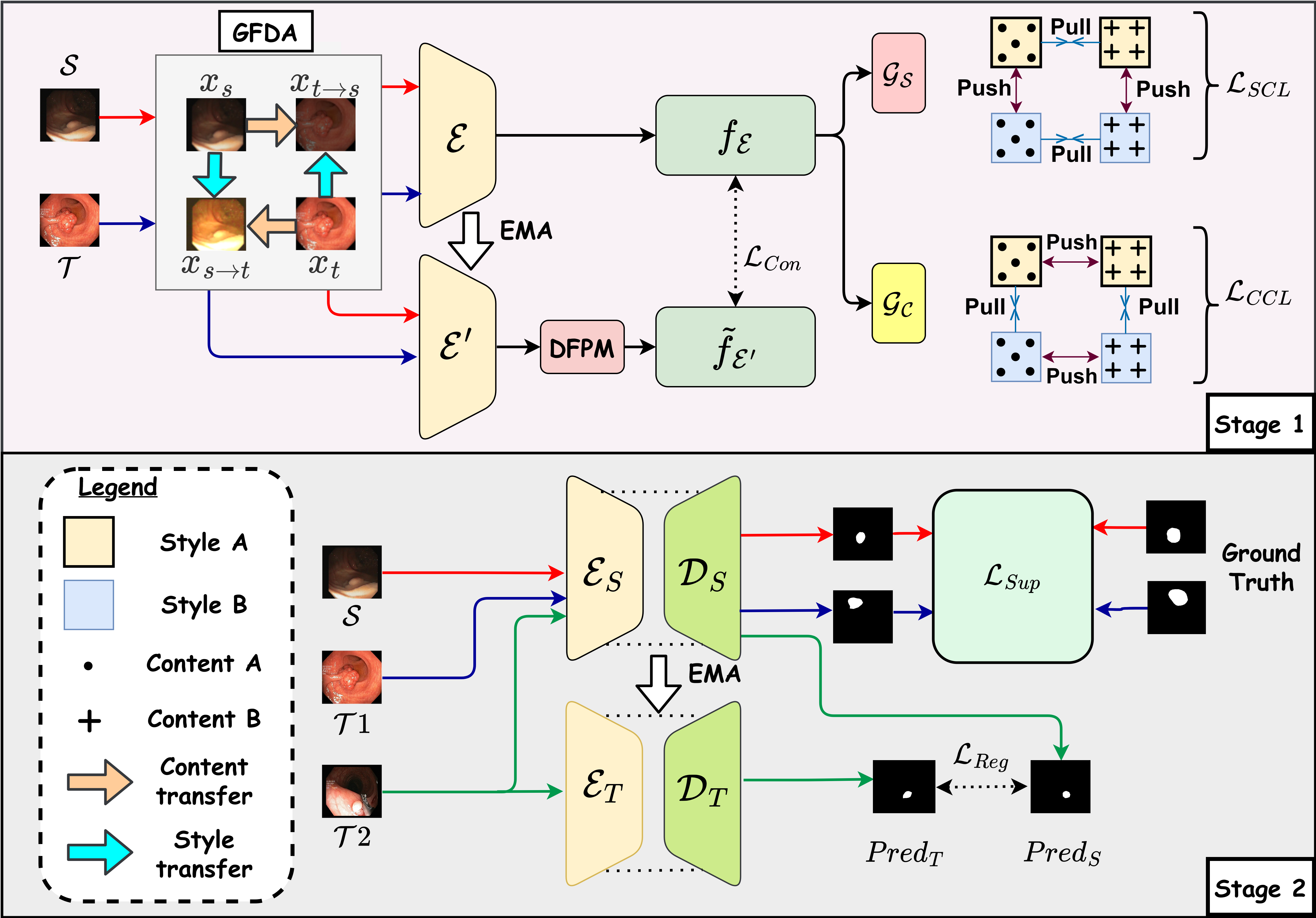}
    \caption{Overall workflow of our proposed method. \textbf{Stage 1}: Encoder pre-training by GFDA and CL on disentangled style and content branches, and pixel-wise feature consistency module DFPM; \textbf{Stage 2}: Fine-tuning the encoder in a semi-supervised student-teacher setting.}
    \label{overall_figure}
\end{figure}

Given the source domain image-label pairs $\{(x_s^i, y_s^i)_{i=1}^{\mathbb{N}_s} \in \mathcal{S}\}$, a few image-label pairs from target domain $\{(x_{t1}^i, y_{t1}^i)_{i=1}^{\mathbb{N}_{t1}} \in \mathcal{T}1\}$, and a large number of unlabeled target images $\{(x_{t2}^i)_{i=1}^{\mathbb{N}_{t2}} \in \mathcal{T}2\}$, our proposed pre-training stage learns from images in $\{\mathcal{S}\cup \mathcal{T}; \mathcal{T}=\mathcal{T}1 \cup \mathcal{T}2\}$ in a self-supervised way, without requiring any labels. The following fine-tuning in SSDA considers image-label pairs in $\{\mathcal{S}\cup \mathcal{T}1\}$ for supervised learning alongside unlabeled images $\mathcal{T}2$ in the target domain for unsupervised prediction consistency. 
Our workflow is shown in \autoref{overall_figure}.

\subsection{Gaussian Fourier Domain Adaptation (GFDA)}\label{FDA}
Manipulating the low-level amplitude spectrum of the frequency domain is the easiest way for style transfer between domains \cite{yang2020fda}, without notable alteration in the visuals of high-level semantics. However, as observed in \cite{yang2020fda}, the generated images consist of incoherent dark patches, caused by abrupt changes in amplitude around the rectangular mask. Instead, we propose a Gaussian mask for a smoother transition in frequency. 
Let, $\mathcal{F}_A(\cdot)$ and $\mathcal{F}_P(\cdot)$ be the amplitude and phase spectrum in frequency space of an RGB image, and $\mathcal{F}^{-1}$ indicates inverse Fourier transform. We define a 2D Gaussian mask $g_{\sigma}$ of the same size as $\mathcal{F}_A$, with $\sigma$ being the standard deviation. Given two randomly sampled images $x_s \sim \mathcal{S}$ and $x_t\sim \mathcal{T}$, our proposed GFDA can be formulated as:
\begin{equation}\label{GFDA_eq}
    x_{s\to t} = \mathcal{F}^{-1}[\mathcal{F}_P(x_s), \mathcal{F}_A(x_t) \odot g_\sigma + \mathcal{F}_A(x_s) \odot (1-g_\sigma) ],
\end{equation}
where $\odot$ indicates element-wise multiplication. It generates an image preserving the semantic content from $\mathcal{S}$ but preserving the style from $\mathcal{T}$. Reciprocal pair $x_{t\to s}$ is also formulated using the same drill. The source and target images, and the style-transferred versions $\{x_s, x_{s\to t}, x_t, x_{t\to s}\}$ are then used for contrastive pre-training below. Visualization of GFDA is shown in the supplementary file.

\subsection{CL on Disentangled Domain and Content}\label{CL}
We aim to learn discriminative content-specific features that are invariant of the style of the source or target domain, for a better pre-training of the network for the task at hand. Hence, we propose to disentangle the style and content information from the images and learn them jointly in a novel disentangled CL paradigm: Style CL (\textit{SCL}) and Content CL (\textit{CCL}). The proposed \textit{SCL} imposes learning of domain-specific attributes, whereas \textit{CCL} enforces the model to identify the ROI, irrespective of the spatial semantics and appearance. In joint learning, they complement each other to render the model to learn domain-agnostic and content-specific information, thereby mitigating the domain dilemma. 
The set of images $\{x_s, x_{s\to t}, x_t, x_{t\to s}\}$, along with their augmented versions are passed through encoder $\mathcal{E}$, followed by two parallel projection heads, namely style head $(\mathcal{G}_\mathcal{S})$ and content head $(\mathcal{G}_\mathcal{C})$ to obtain the corresponding embeddings. Two different losses: style contrastive loss $\mathcal{L}_{SCL}$ and content contrastive loss $\mathcal{L}_{CCL}$, are derived below.  

Assuming $\{x_s,x_{t\to s}\}$ (along with their augmentations) having source-style representation (style $A$), and $\{x_t, x_{s\to t}\}$ (and their augmentations) having target-style representation (style $B$), in style CL, embeddings from the same domain (style) are grouped together whereas embeddings from different domains are pushed apart in the latent space. Considering the $i^{th}$ anchor point $x^i_t\in \mathcal{T}$ in a minibatch and its corresponding style embedding $s^i_t \leftarrow \mathcal{G}_{\mathcal{S}}(\mathcal{E}(x^i_t))$ (with style $B$), we define the \textit{positive} set consisting of the same target domain representations as $\Lambda^+=\{s_t^{j+}, s_{s\to t}^{j+}\}\leftarrow \mathcal{G}_{\mathcal{S}}(\mathcal{E}(\{x_t^j, x_{s\to t}^j\})),\forall j\in \text{minibatch}$, and \textit{negative} set having unalike source domain representation as $\Lambda^-=\{s_s^{j-}, s_{t\to s}^{j-}\}\leftarrow \mathcal{G}_{\mathcal{S}}(\mathcal{E}(\{x_s^j, x_{t\to s}^j\})),\forall j\in\text{minibatch}$. Following SimCLR \cite{chen2020simple} our style contrastive loss can be formulated as:
\begin{equation}
    \mathcal{L}_{SCL} = \sum_{i,j}-\log\frac{\exp(sim(s^i, s^{j+})/\tau)}{\exp(sim(s^i, s^{j+})/\tau) + \sum_{j\in \Lambda^-} \exp(sim(s^i, s^{j-})/\tau)},
\end{equation}
where $\{s^i,s^{j+}\}\in\text{style }B;\;s^{j-}\in\text{style }A$, $sim(\cdot,\cdot)$ defines cosine similarity, $\tau$ is the temperature parameter \cite{chen2020simple}. Similarly, we define $\mathcal{L}_{CCL}$ for content head as:
\begin{equation}
    \mathcal{L}_{CCL} =\sum_{i,j} -\log\frac{\exp(sim(c^i, c^{j+})/\tau)}{\exp(sim(c^i, c^{j+})/\tau) + \sum_{j\in \Lambda^-} \exp(sim(c^i, c^{j-})/\tau)},
\end{equation}
where $\{c^i,c^j\}\leftarrow\mathcal{G}_C(\mathcal{E}(\{x^i,x^j\}))$. These contrastive losses, along with the consistency constraint below enforce the encoder to extract domain-invariant and content-specific feature embeddings. 

\subsection{Consistency Constraint}\label{consistency}
The disentangled CL aims to learn global image-level representation, which is useful for instance discrimination tasks. However, segmentation is attributed to learning dense pixel-level representations. Hence, we propose an additional Dense Feature Propagation Module (DFPM) along with a momentum encoder $\mathcal{E}'$ with exponential moving average (EMA) of parameters from $\mathcal{E}$. 
Given any pixel $m$ of an image $x$, we transform its feature $f^m_{\mathcal{E}'}$ obtained from $\mathcal{E}'$ by  propagating other pixel features from the same image:
\begin{equation}
    \Tilde{f}^m_{\mathcal{E}'} = \sum_{\forall n\in x} \mathcal{K}(f_{\mathcal{E}'}^m)\otimes \cos(f_{\mathcal{E}'}^m, f_{\mathcal{E}'}^n)
\end{equation}
where $\mathcal{K}$ is a linear transformation layer, $\otimes$ denotes \textit{matmul} operation. This spatial smoothing of learned representation is useful for structural sensitivity, which is fundamental for dense segmentation tasks. We enforce consistency between this smoothed feature $\Tilde{f}_{\mathcal{E}'}$ from $\mathcal{E}'$ and the regular feature $f_{\mathcal{E}}$ from $\mathcal{E}$ as:
\begin{equation}
    \mathcal{L}_{Con} = \sum_{[{d}(m,n)<Th]} -\big[\cos(\Tilde{f}^m_{\mathcal{E}'},f^n_\mathcal{E})+\cos(f^m_{\mathcal{E}},\Tilde{f}^n_{\mathcal{E}'})\big]
\end{equation}
where ${d}(\cdot,\cdot)$ indicates the spatial distance, $Th$ is a threshold. The overall pre-training objective can be summarized as:
\begin{equation}
    \mathcal{L}_{Pre} = \lambda_1\mathcal{L}_{SCL}+\lambda_2\mathcal{L}_{CCL}+\mathcal{L}_{Con}
\end{equation}
%
\subsection{Semi-supervised Fine-tuning}\label{fine-tuning}
The pre-training stage is followed by semi-supervised fine-tuning using a student-teacher framework \cite{tarvainen2017mean}. The pre-trained encoder $\mathcal{E}$, along with a decoder $\mathcal{D}$ are used as a student branch, whereas an identical encoder-decoder network (but differently initialized) is used as a teacher network. We compute a supervised loss on the labeled set $\{\mathcal{S}\cup\mathcal{T}1\}$ along with a regularization loss between the prediction of the student and teacher branches on the unlabeled set $\{\mathcal{T}2\}$ as:
\begin{equation}
    \mathcal{L}_{Sup} = \frac{1}{\mathbb{N}_s + \mathbb{N}_{t1}}\sum _{x^i\in \{\mathcal{S}\cup\mathcal{T}1\} }CE\left[\mathcal{D}_S\left(\mathcal{E}_S(x^i)\right),y^i \right]
\end{equation}
\begin{equation}
    \mathcal{L}_{Reg} = \frac{1}{ \mathbb{N}_{t2}}\sum _{x^i\in \{\mathcal{T}2\} }CE\left[\mathcal{D}_S\left(\mathcal{E}_S(x^i)\right), \mathcal{D}_T\left(\mathcal{E}_T(x^i)\right) \right]
\end{equation}
where $CE$ indicates cross-entropy loss, $\mathcal{E}_S$, $\mathcal{D}_S$, $\mathcal{E}_T$, $\mathcal{D}_T$ indicate the student and teacher encoder and decoder networks. The student branch is updated using a consolidated loss $\mathcal{L} = \mathcal{L}_{Sup} + \lambda_3\mathcal{L}_{Reg}$, whereas the teacher parameters $(\theta_T)$ are updated using EMA from the student parameters $(\theta_S)$:
\begin{equation}
    \theta_T (t) = \alpha\theta_T(t-1) + (1-\alpha)\theta_S(t)
\end{equation}
where $t$ tracks the step number, and $\alpha$ is the momentum coefficient \cite{he2020momentum}. 

In summary, the overall SSDA training process contains pre-training (\autoref{FDA}-\autoref{consistency}) and fine-tuning (\autoref{fine-tuning}), whereas, we only use the student branch $(\mathcal{E}_S,\mathcal{D}_S)$ for inference.  

\section{Experiments and Results}\label{Results}
\textbf{Datasets}: We evaluate our work on two different DA tasks to evaluate its generalizability: \textbf{(1)} Polyp segmentation from colonoscopy images in Kvasir-SEG \cite{jha2020kvasir} and CVC-EndoScene Still \cite{vazquez2017benchmark}, and \textbf{(2)} Brain tumor segmentation in MRI images from BraTS2018 \cite{menze2014multimodal}. Kvasir and CVC contain 1000 and 912 images respectively and were split into $4:1$ training-testing sets following \cite{huy2022adversarial}. BraTS consists of brain MRIs from 285 patients with T1, T2, T1CE, and FLAIR scans. The data was split into $4:1$ train-test ratio, following \cite{liu2022act}. 
\textbf{Source$\to$Target}: We perform experiments on $CVC\to Kvasir$ and $Kvasir\to CVC$ for polyp segmentation, and $T2\to \{T1,T1CE,FLAIR\}$ for tumor segmentation. The SSDA accesses $10-50\%$ and $1-5$ labels from the target domain for the two tasks, respectively. For UDA, only $\mathcal{S}$ is used for $\mathcal{L}_{Sup}$, whereas $\mathcal{T}1\cup\mathcal{T}2$ is used for $\mathcal{L}_{Reg}$. \textbf{Implementation details}: Implementation is done in a PyTorch environment using a Tesla V100 GPU with 32GB RAM.  We use U-Net \cite{ronneberger2015u} backbone for the encoder-decoder structure, and the projection heads $\mathcal{G}_\mathcal{S}$ and $\mathcal{G}_\mathcal{C}$ are shallow FC layers. The model is trained for $300$ epochs for pre-training and $500$ epochs for fine-tuning using an ADAM optimizer with a batch size of $4$ and a learning rate of $1e-4$. $\lambda1,\lambda2,\lambda3$, and $Th$ are set to $0.75,0.75,0.5,0.6$, respectively by validation, $\tau, \alpha$ are set to $0.07,0.999$ following \cite{he2020momentum}. Augmentations include random rotation and translation. \textbf{Metrics}: Segmentation performance is evaluated using Dice Similarity Score (DSC) and Hausdorff Distance (HD).
\begin{table}[!t]
\centering
\resizebox{0.65\textwidth}{!}{%
\begin{tabular}{@{}crcccccc@{}}
\toprule
{ }                     &             & { }                                & \multicolumn{2}{c}{{ \textbf{CVC $\rightarrow$  Kvasir}}}        &   & \multicolumn{2}{c}{{ \textbf{Kvasir $\rightarrow$ CVC}}}             \\ \cmidrule(l){4-8} 
\multirow{-2}{*}{{ \textbf{Task}}} & \multirow{-2}{*}{{ \textbf{Method}}} & \multirow{-2}{*}{{ \textbf{Target label}}} & { \textbf{DSC}}$\uparrow$  & { \textbf{HD}}$\downarrow$  & &{ \textbf{DSC}}$\uparrow$  & { \textbf{HD}}$\downarrow$  \\ \midrule
No DA & { Source only}                       & { $0\%$L}                           & { 62.2}          & { 5.6}    &      & { 53.9}          & { 6.2}          \\ \midrule
\multirow{6}{*}{{ {UDA}}} & { PCEDA} \cite{yang2020phase}                             & { $0\%$L}                             & { 73.6}          & { 4.4}     &     & { 70.1}          & { 4.7}          \\
&{ ASN} \cite{tsai2018learning}                              & { $0\%$L}                             & { 80.1}          & {\color[HTML]{3531FF} 3.6}    &      & {\color[HTML]{3531FF} 83.7}          & { 3.7}          \\
&{ BDL} \cite{li2019bidirectional}                              & { $0\%$L}                             & { 77.8}          & { 4.0}    &      & { 81.7}          & { 4.1}          \\
&{ CoFo} \cite{huy2022adversarial}                             & { $0\%$L}                             & {\color[HTML]{3531FF} 82.8}          & {\color[HTML]{3531FF} 3.6}    &      & { 81.1}          & {\color[HTML]{3531FF} 3.5}          \\
&{ FDA} \cite{yang2020fda}                               & { $0\%$L}                             & { 80.4}          & { 3.9}    &      & { 75.1}          & { 4.2}          \\
&{ \textbf{Ours}}                     & { \textbf{0\%L}}                    & {\color[HTML]{FE0000} \textbf{83.8}} & {\color[HTML]{FE0000} \textbf{3.4}} & & {\color[HTML]{FE0000} \textbf{84.5}} & {\color[HTML]{FE0000} \textbf{3.1}} \\ \midrule
\multirow{10}{*}{{ {SSDA}}}&{ DLD} \cite{wang2020alleviating}                              & { 10\%L}                     & { 84.2}          & { 3.2}     &     & { 85.1}          & { 3.1}          \\
&{ ACT} \cite{liu2022act}                              & { 10\%L}                     & {\color[HTML]{3531FF} 86.9}          & {\color[HTML]{3531FF} 3.0}    &      & {\color[HTML]{FE0000} 87.3}          & {\color[HTML]{3531FF} 2.9}          \\
&{ SLA} \cite{chen2021semi}                              & { 10\%L}                     & { 85.5}          & { 3.1}     &     & { 86.2}          & { 3.3}          \\
&{ FSM} \cite{yang2022source}                              & { 10\%L}                     & { 85.8}          & { 3.4}     &     & { 86.2}          & { 3.1}          \\
&{ \textbf{Ours}}                     & { \textbf{10\%L}}            & {\color[HTML]{FE0000} \textbf{87.7}} & {\color[HTML]{FE0000} \textbf{2.9}} & & {\color[HTML]{3531FF} \textbf{86.9}} & {\color[HTML]{FE0000} \textbf{2.7}} \\ \cmidrule(l){2-8}
&{ DLD} \cite{wang2020alleviating}                              & { 50\%L}                     & { 87.6}          & { 2.8}    &      & { 87.9}          & { 2.6}          \\
&{ ACT} \cite{liu2022act}                              & { 50\%L}                     & {\color[HTML]{3531FF} 89.4}          & {\color[HTML]{3531FF} 2.6}    &      & {\color[HTML]{3531FF} 90.3}          & {\color[HTML]{3531FF} 2.4}          \\
&{ SLA} \cite{chen2021semi}                              & { 50\%L}                     & { 88.6}          & { 2.7}    &      & { 89.3}          & { 2.8}          \\
&{ FSM} \cite{yang2022source}                              & { 50\%L}                     & { 89.1}          & {\color[HTML]{3531FF} 2.6}    &      & { 89.8}          & { 2.5}          \\
&{ \textbf{Ours}}                     & { \textbf{50\%L}}            & {\color[HTML]{FE0000} \textbf{90.6}} & {\color[HTML]{FE0000} \textbf{2.4}}  & & {\color[HTML]{FE0000} \textbf{90.8}} & {\color[HTML]{FE0000} \textbf{2.2}} \\ \midrule
Supervised & { Source+Target}                     & { 100\%L}                      & { 92.1}          & { 2.1}     &     & { 93.8}          & { 2.0}          \\ \bottomrule
\end{tabular}%
}
\caption{Comparison with state-of-the-art UDA and SSDA methods for polyp segmentation on KVASIR and CVC. SSDA results are shown for 10\%-labeled (10\%L) and 50\%-labeled (50\%L) data in the target domain. The results of cited methods are directly reported from the corresponding papers. \textbf{No DA}: the encoder-decoder model trained only using labeled data from the source domain is applied to the target domain without adaptation. \textbf{Supervised}: model is trained using all labeled data from source and target domains. The best and second-best results are highlighted in \textcolor{red}{RED} and \textcolor{blue}{BLUE}, respectively.}
\label{tab:comparison-polyp}
\end{table}

\begin{table}[!tbp]
\centering
\resizebox{0.65\textwidth}{!}{%
\begin{tabular}{@{}crccccccc@{}}
\toprule
                          &        &                                 & \multicolumn{3}{c}{\textbf{DSC}$\uparrow$}                                                                      & \multicolumn{3}{c}{\textbf{HD}$\downarrow$}                                                              \\ \cmidrule(l){4-9} 
\multirow{-2}{*}{\textbf{Task}} & \multirow{-2}{*}{\textbf{Method}} & \multirow{-2}{*}{\textbf{Target Label}} & \textbf{T1}                          & \textbf{T1CE}                        & \textbf{FLAIR}                       & \textbf{T1}                          & \textbf{T1CE}                        & \textbf{FLAIR}                      \\ \midrule
No DA & Source only                       & $0$L                           & 3.9                                  & 6.0                                  & 64.4                                 & 56.9                                 & 50.8                                 & 30.4                                \\ \midrule
\multirow{5}{*}{{ {UDA}}} & SSCA \cite{liu2022self}                             & $0$L                             & {\color[HTML]{3531FF} 59.3}          & {\color[HTML]{3531FF} 63.5}          & {\color[HTML]{3531FF} 82.9}          & {\color[HTML]{3531FF} 12.5}          & {\color[HTML]{3531FF} 11.2}          & {\color[HTML]{3531FF} 7.9}          \\
& SIFA \cite{chen2019synergistic}                             & $0$L                             & 51.7                                 & 58.2                                 & 68.0                                 & 19.6                                 & 15.0                                 & 16.9                                \\
& DSA \cite{han2022deep}                              & $0$L                             & 57.7                                 & 62.0                                 & 81.8                                 & 14.2                                 & 13.7                                 & 8.6                                 \\
& DSFN \cite{zou2022unsupervised}                             & $0$L                             & 57.3                                 & 62.2                                 & 78.9                                 & 17.5                                 & 15.5                                 & 13.8                                \\
& \textbf{Ours}                     & \textbf{0L}                    & {\color[HTML]{FE0000} \textbf{60.7}} & {\color[HTML]{FE0000} \textbf{64.4}} & {\color[HTML]{FE0000} \textbf{83.3}} & {\color[HTML]{FE0000} \textbf{11.1}} & {\color[HTML]{FE0000} \textbf{10.9}} & {\color[HTML]{FE0000} \textbf{7.3}} \\ \midrule
\multirow{10}{*}{{ {SSDA}}} & DLD \cite{wang2020alleviating}                              & 1L                        & 65.8                                 & 66.5                                 & 81.5                                 & 12.0                                 & 10.3                                 & 7.1                                 \\
& ACT \cite{liu2022act}                              & 1L                        & {\color[HTML]{3531FF} 69.7}          & {\color[HTML]{3531FF} 69.7}          & {\color[HTML]{3531FF} 84.5}          & {\color[HTML]{3531FF} 10.5}          & {\color[HTML]{3531FF} 10.0}          & {\color[HTML]{3531FF} 5.8}          \\
& ACT-EMD \cite{liu2022act}                              & 1L                        &  67.4          & 69.0          & 83.9          & 10.9          & 10.3          & 6.4          \\

& SLA \cite{chen2021semi}                              & 1L                        & 64.7                                 & 66.1                                 & 82.3                                 & 12.2                                 & 10.5                                 & 7.1                                 \\
& \textbf{Ours}                     & \textbf{1L}               & {\color[HTML]{FE0000} \textbf{72.2}} & {\color[HTML]{FE0000} \textbf{71.9}} & {\color[HTML]{FE0000} \textbf{85.8}} & {\color[HTML]{FE0000} \textbf{10.0}} & {\color[HTML]{FE0000} \textbf{9.5}}  & {\color[HTML]{FE0000} \textbf{5.2}} \\ \cmidrule(l){2-9} 
& DLD \cite{wang2020alleviating}                              & 5L                        & 67.8                                 & 68.3                                 & 83.3                                 & 11.2                                 & 9.9                                  & 6.6                                 \\
& ACT \cite{liu2022act}                              & 5L                        & {\color[HTML]{3531FF} 71.3}          & 70.8                                 & {\color[HTML]{3531FF} 85.0}          & {\color[HTML]{3531FF} 10.0}          & {\color[HTML]{3531FF} 9.8}           & {\color[HTML]{3531FF} 5.2}          \\
& ACT-EMD \cite{liu2022act}                              & 5L                        & 70.3          & 69.8                                 & 84.4          & 10.4          & 10.2           & 5.7          \\
& SLA \cite{chen2021semi}                               & 5L                        & 67.2                                 & {\color[HTML]{3531FF} 71.2}          & 83.1                                 & 11.7                                 & 10.1                                 & 6.8                                 \\
& \textbf{Ours}                     & \textbf{5L}               & {\color[HTML]{FE0000} \textbf{73.1}} & {\color[HTML]{FE0000} \textbf{72.4}} & {\color[HTML]{FE0000} \textbf{86.1}} & {\color[HTML]{FE0000} \textbf{9.7}}  & {\color[HTML]{FE0000} \textbf{9.3}}  & {\color[HTML]{FE0000} \textbf{4.8}} \\ \midrule
Supervised & Source+Target          & all labeled                 & 73.6                                 & 72.9                                 & 86.6                                 & 9.5                                  & 9.1                                  & 4.6                                 \\ \bottomrule
\end{tabular}%
}
\caption{Comparison with state-of-the-art UDA and SSDA methods for whole tumor segmentation on BraTS2018, where source domain is T2. SSDA results are demonstrated for 1-labeled (1L) and 5-labeled (5L) data in the target domain. }
\label{tab:comparison-brats}
\end{table}

\subsection{Performance on SSDA}\label{SSDA_result}
Quantitative comparison of our proposed method with different SSDA methods \cite{wang2020alleviating,liu2022act,chen2021semi,yang2022source} for both tasks are shown in \autoref{tab:comparison-polyp} and \autoref{tab:comparison-brats}. 
ACT \cite{liu2022act} simply ignores the domain gap and only learns content semantics, resulting in substandard performance on the BraTS dataset that has a significant domain gap. FSM \cite{yang2022source}, on the other hand, is adaptable to learning explicit domain information, but lacks strong pixel-level regularization on its prediction, resulting in subpar performance. We address both of these shortcomings in our work, resulting in superior performance on both tasks. Other methods like \cite{wang2020alleviating,chen2021semi}, which are originally designed for natural images, lack critical refining abilities even after fine-tuning for medical image segmentation and hence are far behind our performance in both tasks. The margins are even higher for less labeled data (1L) on the BraTS dataset, which is promising considering the difficulty of the task. Moreover, our method produces performance close to its fully-supervised counterpart (last row in \autoref{tab:comparison-polyp} and \autoref{tab:comparison-brats}), using only a few target labels.  

\subsection{Performance on UDA}\label{UDA_result}
Unlike SSDA methods, UDA fully relies on unlabeled data for domain-invariant representation learning. To analyze the effectiveness of DA, we extend our model to the UDA setting (explained in \autoref{Results}[\textbf{Source}$\to$\textbf{Target}]) and compare it with SoTA methods \cite{liu2022self,chen2019synergistic,huy2022adversarial,yang2020fda,zou2022unsupervised,han2022deep,yang2020phase,li2019bidirectional} in \autoref{tab:comparison-polyp} and \autoref{tab:comparison-brats}. Methods like \cite{huy2022adversarial,tsai2018learning} rely on adversarial learning for aligning multi-level feature space, which is not effective for small-sized medical data. Other methods \cite{yang2020phase,li2019bidirectional} rely on an image-translation network but fail in effective style adaptation, resulting in source domain-biased subpar performance. Our method, although relies on FDA \cite{yang2020fda}, outperforms it with a large margin of upto $12.5\%$ DSC for polyp segmentation, owing to its superior learning ability of disentangled style and content semantics. Similar results are observed for the BraTS dataset in \autoref{tab:comparison-brats}, where our work achieved a margin of upto $2.4\%$ DSC than its closest performer.      

\subsection{Ablation Experiments}\label{ablation}
We perform a detailed ablation experiment, as shown in \autoref{tab:ablation}. The effectiveness of disentangling and joint-learning of style and content information is evident from the experiment (b)\&(c) as compared to (a), where the introduction of \textit{SCL} and \textit{CCL} boosts overall performance significantly. Moreover, when combined together (experiment (d)), they provide a massive $9.54\%$ and $8.52\%$ DSC gain over traditional CL (experiment (a)) for $CVC\to Kvasir$ and $Kvasir\to CVC$, respectively. This also points out a potential shortfall of traditional CL: its inability to adapt to a complex domain in DA. The proposed DFPM (experiment (e))  provides local pixel-level regularization, complementary to the global disentangled CL, resulting in a further boost in performance ($\sim 1.5\%$). We have similar ablation study observations on the BraTS2018 dataset, which is provided in the supplementary file, along with some qualitative examples along with available ground truth.

\begin{table}[!t]
\centering
\resizebox{0.8\textwidth}{!}{%
\begin{tabular}{@{}cccccccccccc@{}}
\toprule
\multirow{2}{*}{Experiment\#} & \multicolumn{4}{c}{\textbf{Stage 1}}                                                                                                                      & \textbf{Stage 2}   &                      & \multicolumn{2}{c}{CVC $\rightarrow$ Kvasir}                                             &                                  & \multicolumn{2}{c}{Kvasir $\rightarrow$ CVC}                                             \\ \cmidrule(l){2-12}
& \textbf{TCL}                           & \textbf{SCL}                            & \textbf{CCL}                          & \textbf{DFPM}                                & \textbf{SemiSL}     &                         & \textbf{DSC}$\uparrow$                                  & \textbf{HD}$\downarrow$                                  &                                  & \textbf{DSC}$\uparrow$                                  & \textbf{HD}$\downarrow$                                  \\ \midrule
(a) & $\checkmark$                                & $\times$                                  & $\times$                                  & $\times$                                  & $\checkmark$                &                 & 81.7                                 & 4.4                                 &                                  & 82.1                                 & 4.2                                 \\
(b) & $\times$                                 & $\checkmark$                                 & $\times$                                  & $\times$                                  & $\checkmark$            &                     & 83.2                                 & 3.9                                 &                                  & 84.7                                 & 3.5                                 \\
(c) & $\times$                                 & $\times$                                  & $\checkmark$                                 & $\times$                                  & $\checkmark$            &                     & 84.5                                 & 3.8                                 &                                  & 85.4                                 & 3.1                                 \\
(d) & $\times$                                 & $\checkmark$                                 & $\checkmark$                                 & $\times$                                  & $\checkmark$          &                       & 89.5                                 & 2.8                                 &                                  & 89.1                                 & 2.4                                 \\
\textcolor{red}{(e)} & {\color[HTML]{FE0000} \textbf{$\times$}} & {\color[HTML]{FE0000} \textbf{$\checkmark$}} & {\color[HTML]{FE0000} \textbf{$\checkmark$}} & {\color[HTML]{FE0000} \textbf{$\checkmark$}} & {\color[HTML]{FE0000} \textbf{$\checkmark$}} & & {\color[HTML]{FE0000} \textbf{90.6}} & {\color[HTML]{FE0000} \textbf{2.4}} & {\color[HTML]{FE0000} \textbf{}} & {\color[HTML]{FE0000} \textbf{90.8}} & {\color[HTML]{FE0000} \textbf{2.2}} \\ \bottomrule
\end{tabular}%
}
\caption{Ablation experiment for polyp segmentation in SSDA(50\%L) setting to identify the contribution of individual components. TCL: traditional CL \cite{chaitanya2020contrastive}, SCL: proposed style CL, CCL: proposed content CL. The last row, highlighted in \textcolor{red}{\textbf{RED}}, indicates our results.}
\label{tab:ablation}
\end{table}

\section{Conclusion}\label{Conclusion}
We propose a novel style-content disentangled contrastive learning, guided by a pixel-level feature consistency constraint for semi-supervised domain adaptive medical image segmentation. To the best of our knowledge, this is the first attempt for SSDA in medical image segmentation using CL, which is further extended to the UDA setting.  Our proposed work, upon evaluation on two different domain adaptive segmentation tasks in SSDA and UDA settings, outperforms the existing SoTA methods, justifying its effectiveness and generalizability.

\title{Supplementary File: Semi-supervised Domain Adaptive Medical Image Segmentation through Consistency Regularized Disentangled Contrastive Learning}
\titlerunning{Supplementary for Disentangled Consistency Contrast}
%
\author{Hritam Basak, Zhaozheng Yin} 
%
\authorrunning{H. Basak et al.}
%
\institute{Dept. of Computer Science, Stony Brook University, NY, USA}
%
\maketitle              

\begin{figure}[!h]
    \centering
    \includegraphics[width=0.95\columnwidth]{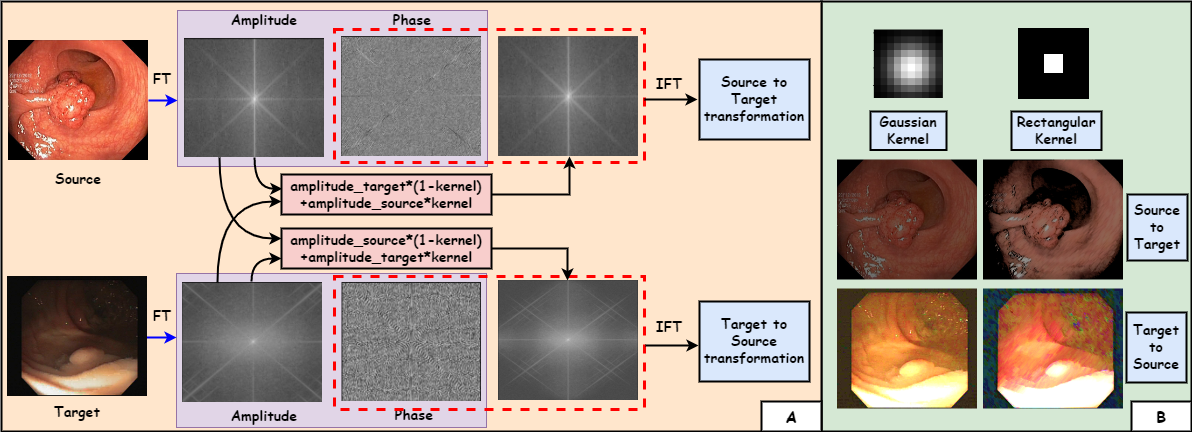}
    \caption{Visualization of our proposed GFDA module, FT: Fourier Transform, IFT: Inverse Fourier Transform; \textbf{(A)} The Gaussian spectral transfer method of changing image \textit{style} without altering semantic \textit{content} information; \textbf{(B)} Qualitative comparison of our proposed method along with traditional FDA method with a fixed rectangular kernel. Clearly, GFDA results in smoother and noise-free intensity transitions in the reconstructed images.}
    \label{fig:GFDA}
\end{figure}

\begin{figure}[!h]
    \centering
    \includegraphics[width=0.95\columnwidth, height = 58mm]{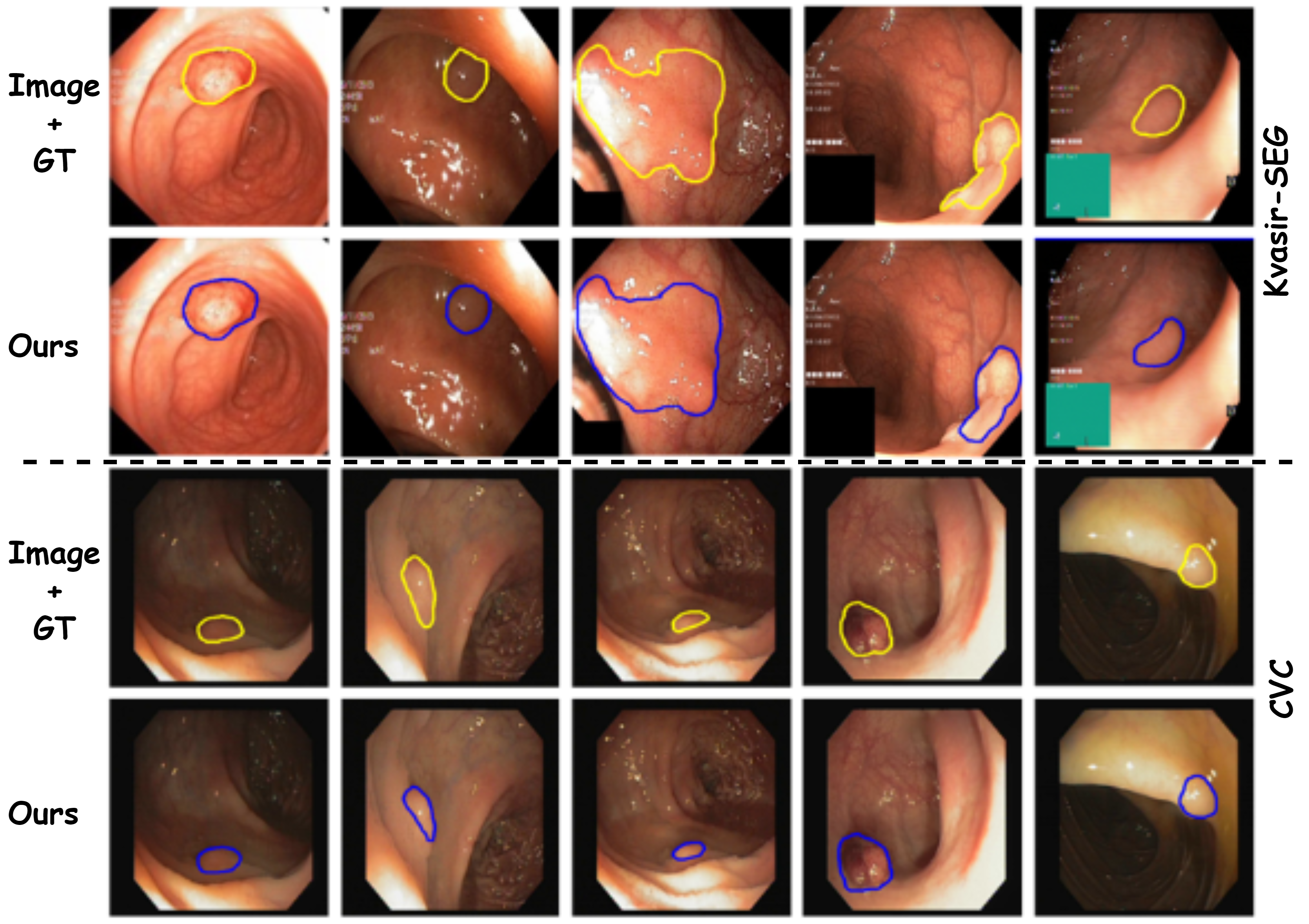}
    \caption{Qualitative analysis of our segmentation performance for polyp segmentation on target-domain Kvasir-SEG and CVC datasets.}
    \label{fig:qualitative_polyp}
\end{figure}
\begin{figure}[!h]
    \centering
    \includegraphics[width = \columnwidth, height = 100mm]{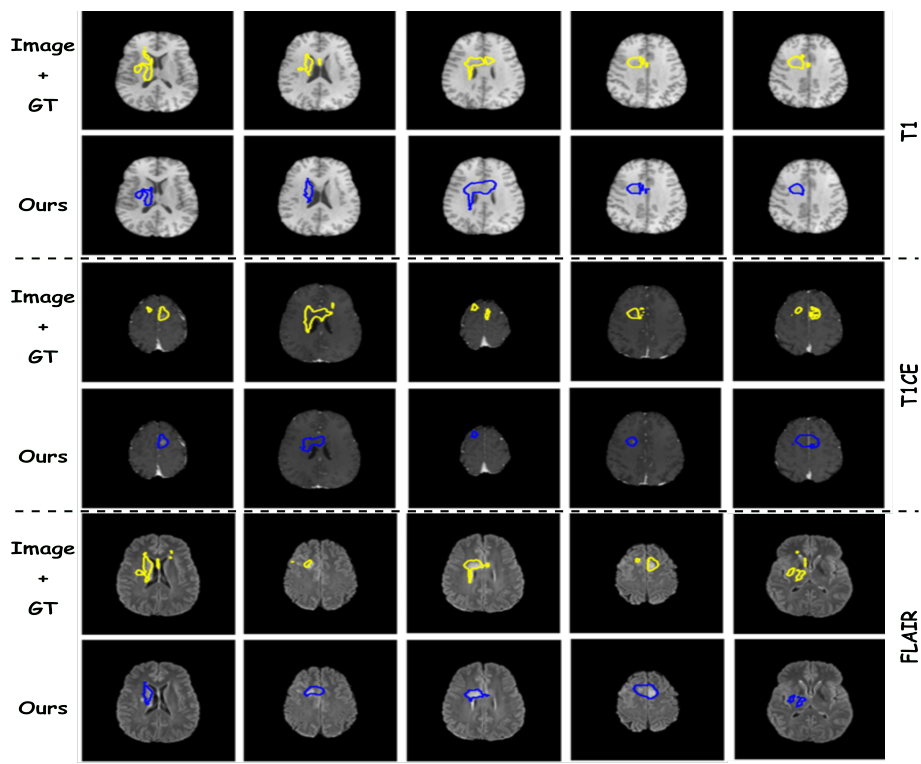}
    \caption{Qualitative analysis of our segmentation performance for tumor segmentation from BraTS2018 dataset on target-domain T1, T1CE, and FLAIR modalities, where T2 is used as source-domain.}
    \label{fig:qualitative_brats}
\end{figure}

\begin{table}[!h]
\centering
\resizebox{0.9\textwidth}{!}{%
\begin{tabular}{@{}cccccccccccccc@{}}
\toprule
                                         & \multicolumn{4}{c}{\textbf{Stage 1}}                                                                                                             & \textbf{Stage 2}                & \multicolumn{2}{c}{\textbf{T2$\to$T1}}                                         & \textbf{}                        & \multicolumn{2}{c}{\textbf{T2$\to$T1CE}}                                       & \textbf{}               & \multicolumn{2}{c}{\textbf{T2$\to$FLAIR}}                                      \\ \cmidrule(l){2-14} 
\multirow{-2}{*}{\textbf{Experiment \#}} & \textbf{TCL}                  & \textbf{SCL}                   & \textbf{CCL}                 & \textbf{DFPM}                       & \textbf{SemiSL}                     & \textbf{DSC}$\uparrow$                         & \textbf{HD}$\downarrow$                         &                                  & \textbf{DSC}$\uparrow$                         & \textbf{HD}$\downarrow$                         &                         & \textbf{DSC}$\uparrow$                         & \textbf{HD}$\downarrow$                        \\ \midrule
(a)                                      & $\checkmark$                                & $\times$                                  & $\times$                                  & $\times$                                  & $\checkmark$                                 & 63.6                                 & 11.6                                &                                  & 62.8                                 & 11.8                                &                         & 77.6                                 & 8.8                                 \\
(b)                                      & $\times$                                 & $\checkmark$                                 & $\times$                                  & $\times$                                  & $\checkmark$                                & 67.4                                 & 10.7                                &                                  & 66.4                                 & 10.4                                &                         & 80.3                                 & 7.9                                 \\
(c)                                      & $\times$                                 & $\times$                                  & $\checkmark$                                 & $\times$                                  & $\checkmark$                                 & 67.8                                 & 10.6                                &                                  & 67.7                                 & 10.3                                &                         & 80.9                                 & 7.7                                 \\
(d)                                      & $\times$                                 & $\checkmark$                                 & $\checkmark$                                 & $\times$                                  & $\checkmark$                                 & 72.2                                 & 10.0                                &                                  & 71.7                                 & 9.7                                 &                         & 85.3                                 & 5.1                                 \\
{\color[HTML]{FE0000} \textbf{(e)}}      & {\color[HTML]{FE0000} \textbf{$\times$}} & {\color[HTML]{FE0000} \textbf{$\checkmark$}} & {\color[HTML]{FE0000} \textbf{$\checkmark$}} & {\color[HTML]{FE0000} \textbf{$\checkmark$}} & {\color[HTML]{FE0000} \textbf{$\checkmark$}} & {\color[HTML]{FE0000} \textbf{73.1}} & {\color[HTML]{FE0000} \textbf{9.7}} & {\color[HTML]{FE0000} \textbf{}} & {\color[HTML]{FE0000} \textbf{72.4}} & {\color[HTML]{FE0000} \textbf{9.3}} & {\color[HTML]{FE0000} } & {\color[HTML]{FE0000} \textbf{86.1}} & {\color[HTML]{FE0000} \textbf{4.8}} \\ \bottomrule
\end{tabular}%
}
\caption{Ablation experiment for tumor segmentation on BraTS2018 dataset in SSDA(5L) setting to identify the contribution of individual components. TCL: traditional CL [\textcolor{red}{2}], SCL: proposed style CL, CCL: proposed content CL. The last row, highlighted in \textcolor{red}{RED}, indicates our results. }
\label{tab:ablation-brats}
\end{table}

%



\bibliographystyle{splncs04}
\bibliography{reference}
\end{document}